\documentclass[sigconf]{acmart}
\AtBeginDocument{%
  }

\copyrightyear{2026}
\acmYear{2026}
\setcopyright{cc}
\setcctype{by-nc-nd}
\acmConference[MM '26]{Proceedings of the 34th ACM International Conference on Multimedia}{November 10--14, 2026}{Rio de Janeiro, Brazil}
\acmBooktitle{Proceedings of the 34th ACM International Conference on Multimedia (MM '26), November 10--14, 2026, Rio de Janeiro, Brazil}
\acmDOI{10.1145/3767308.3835891}
\acmISBN{979-8-4007-2213-4/2026/11}

\usepackage{balance}

\begin{document}

\title{TeaMatch: Teachable Cross-Modal Representation Learning for 2D–3D Matching}


\author{Chongjian Wang}
\orcid{0009-0008-7204-3225}

\affiliation{%
  \institution{Shandong Women's University}
  \city{Jinan}
  \country{China}
}

\affiliation{%
  \institution{Shandong University of Science and Technology}
  \city{Qingdao}
  \country{China}
}

\email{202311080223@sdust.edu.cn}

\author{Junjie Gao}
\authornote{Corresponding author.}
\orcid{0000-0001-7087-0886}

\affiliation{%
  \institution{Shandong Women's University}
  \city{Jinan}
  \country{China}
}

\email{junjie.gao@sdwu.edu.cn}

\renewcommand{\shortauthors}{Chongjian Wang and Junjie Gao}


\begin{abstract}
Learning reliable correspondences between images and point clouds is fundamental for 2D–3D matching. Despite recent progress in detection-free methods, existing approaches primarily optimize matching within a single model, and often struggle to maintain reliable correspondences under challenging conditions such as noisy inputs, low overlap, and ambiguous structures. In this work, we propose TeaMatch, a novel framework that introduces teachability as a criterion for cross-modal representation learning. We define teachability as the ability of a representation to be effectively recovered by weak learners under degraded inputs, reflecting its structural consistency and robustness. To this end, we construct a set of task-specific weak students that simulate common failure modes, and train them to imitate the teacher on a training split while evaluating their recoverability on a disjoint meta split. The teacher is then optimized to improve the students’ ability to recover reliable correspondences, guided by correspondence-level and geometry-aware constraints. Our framework can be seamlessly integrated into existing coarse-to-fine matching pipelines without additional inference cost.Extensive experiments demonstrate that TeaMatch improves matching robustness and achieves state-of-the-art performance on challenging 2D–3D matching benchmarks.
\end{abstract}




\maketitle

\section{Introduction}

Establishing reliable correspondences between images and point clouds is fundamental to visual localization~\cite{Selvaraju2016GradCAMVE,Zhou2024TheNM}, robot navigation~\cite{Huang2022VisualLM,Patle2019ARO}, augmented reality~\cite{Azuma1997ASO}, simultaneous localization and mapping~\cite{DurrantWhyte2006SimultaneousLA}, and 3D reconstruction~\cite{Hong2023LRMLR,Leroy2024GroundingIM}. Given an image and a point cloud capturing the same scene, 2D--3D matching aims to identify geometrically consistent pixel--point correspondences, from which the relative pose can be estimated using a robust solver such as PnP-RANSAC~\cite{Fischler1981RandomSC,lepetit2009epnp}. As in other registration problems, correspondence quality is therefore critical to successful downstream localization.

\begin{figure}[t]
\vspace{4pt}
  \centering
  \includegraphics[width=\linewidth]{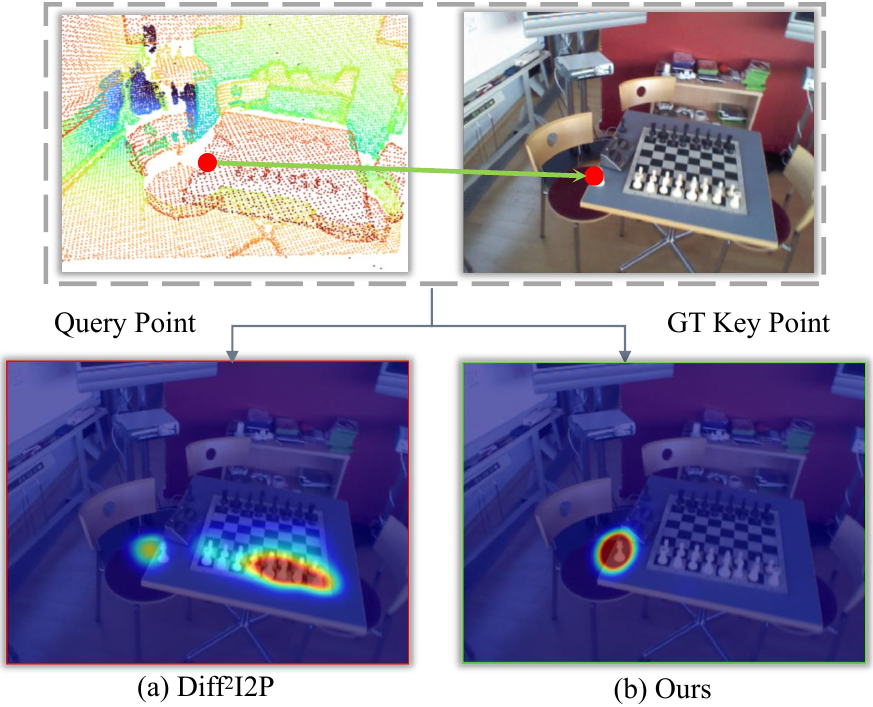}
  \caption{Qualitative comparison of correspondence localization. We select a query point and visualize the matching confidence maps of Diff2I2P and our method. While the baseline assigns high confidence to a broad ambiguous region, TeaMatch localizes the true correspondence with a sharp peak.}
  \label{fig:1}

\end{figure}

Traditional 2D--3D matching pipelines~\cite{Feng20192D3DMatchnetLT,Pham2019LCDLC,Wang2021P2NetJD} typically follow a detect-then-match strategy. They independently detect salient keypoints in the image and point cloud, extract modality-specific descriptors, and establish cross-modal correspondences. However, the intrinsic gap between 2D appearance and 3D geometry makes repeatable keypoints difficult to define consistently across modalities, while descriptors learned from texture-rich images and sparsely sampled point clouds often remain poorly aligned. Consequently, these methods tend to produce low inlier ratios and limited robustness under viewpoint changes, noise, and partial overlap.

Recently, detection-free methods~\cite{Li2021DeepI2PIC,Li20232D3DMATR2M,Ren2022CorrI2PDI,Zhou2023DifferentiableRO,mu2025diff2i2p} have emerged as a strong alternative, directly inferring correspondences from dense or semi-dense cross-modal representations without explicit keypoint detection. Many further adopt coarse-to-fine strategies~\cite{Li20232D3DMATR2M,mu2025diff2i2p}, progressively refining region-level alignments into pixel--point correspondences. Although these frameworks significantly improve robustness, they are typically trained within a single-model paradigm that optimizes a high-capacity matcher without explicitly enforcing robust and recoverable representations. Under challenging conditions such as noise, low overlap, and repetitive structures, models may therefore rely on fragile pair-level representations that are sensitive to corruption and distribution shifts, degrading correspondence quality and downstream pose estimation. This suggests that matching accuracy alone is insufficient to characterize representation quality: beyond being discriminative, cross-modal representations should preserve stable geometric structure under constrained conditions. As illustrated in Fig.~\ref{fig:1}, existing methods often produce diffuse responses over ambiguous regions, whereas reliable representations should yield sharp and well-localized correspondences.

To address this limitation, we propose TeaMatch, a novel teachability-driven framework for cross-modal representation learning in 2D--3D matching. Our key idea is to evaluate representation quality not only through the performance of a strong matcher, but also by whether the learned image--point representations remain recoverable under degraded inputs and constrained learners. Intuitively, if a representation captures stable geometric correspondence structure between 2D appearance and 3D geometry, even a deliberately weak learner should recover reliable correspondence preferences from it.

TeaMatch introduces task-specific weak students that simulate common failure modes and interact with the teacher during training. The students imitate the teacher on one subset of local regions and are evaluated on disjoint unseen regions, encouraging the teacher to produce representations that generalize beyond local redundancy. The teacher is then optimized to improve student recoverability, making the learned representations both discriminative and robust to perturbations and distribution shifts. We further incorporate correspondence-level and geometry-aware constraints to align recoverability with downstream localization, promoting reliable inlier selection and pose estimation.

A key advantage of TeaMatch is that it is a \emph{training-only} framework. All weak learners and teachability-related objectives are removed at inference time. TeaMatch can therefore be integrated into existing detection-free coarse-to-fine pipelines without modifying the inference procedure or introducing additional computational cost. Extensive experiments on challenging 2D--3D matching benchmarks demonstrate that TeaMatch significantly improves matching robustness and achieves state-of-the-art performance.

Our contributions are summarized as follows:
\begin{itemize}
    \item We revisit 2D--3D matching from the perspective of cross-modal representation learning and introduce \emph{teachability} as a criterion for assessing whether pair-level image--point representations are robust, recoverable, and useful for downstream localization.
    \item We propose \textbf{TeaMatch}, a training-only framework that improves representation quality through degraded weak learners, a patch-disjoint teach/meta split, and correspondence- and geometry-aware recoverability supervision.
    \item Extensive experiments show that TeaMatch consistently improves strong detection-free baselines, yielding more robust correspondences and better pose estimation without changing the inference pipeline or adding test-time cost.
\end{itemize}

\section{Related Work}

\subsection{Image and Point Cloud Registration}

Image registration has been extensively studied in computer vision. Classical methods follow a detect-then-match paradigm, where keypoints are described using hand-crafted~\cite{Lowe1999ObjectRF,Rublee2011ORBAE} or learned features~\cite{DeTone2017SuperPointSI,Dusmanu2019D2NetAT,Luo2020ASLFeatLL,Sarlin2019SuperGlueLF}, matched by descriptor similarity, and used to recover the transformation through PnP or bundle adjustment~\cite{Triggs1999BundleA}. Because keypoint detection often degrades under viewpoint, illumination, and texture changes, recent methods~\cite{Lee2021PatchMatchBasedNC,Li2020CorrespondenceNW,Rocco2018NeighbourhoodCN,Rocco2020EfficientNC,Sun2021LoFTRDL,Zhou2020Patch2PixEP} adopt detector-free formulations that infer correspondences directly from dense features. Coarse-to-fine strategies~\cite{Sun2021LoFTRDL} further improve efficiency and localization accuracy by progressively refining global matches into local correspondences.

Point cloud registration has similarly evolved from hand-crafted descriptors~\cite{Rusu2009FastPF,Guo2013RoPSAL,Salti2014SHOTUS,Dong2017ANB}, which are sensitive to noise, density variation, and partial overlap, to learned representations~\cite{Yew20183DFeatNetWS,Deng2018PPFNetGC,Gojcic2018ThePM,Ao2020SpinNetLA} with larger receptive fields. Detector-free pipelines~\cite{Choy2019FullyCG,Thomas2019KPConvFA}, hierarchical matching~\cite{Yu2021CoFiNetRC}, and context-aware interaction~\cite{Gao2023OAAFormerRA,Qin2023GeoTransformerFA,Yu2023RotationInvariantTF,Chen2024DynamicCT,Yao2024PARENetPR} further improve correspondence estimation. Robust estimators~\cite{Bai2021PointDSCRP,Chen2022SC2PCRAS,Jiang2023RobustOR,Xing2024EfficientSC} also extend RANSAC-style methods~\cite{Li2020GESACRG,Fischler1981RandomSC} by explicitly modeling correspondence consistency.

\subsection{Cross-Modal 2D--3D Matching}

Image--point cloud matching is more challenging than single-modality matching because of the gap between 2D appearance and 3D geometry. Early methods~\cite{Feng20192D3DMatchnetLT,Pham2019LCDLC,Wang2021P2NetJD} independently detect keypoints in each modality and match them using cross-modal descriptors. However, repeatable keypoints are difficult to define across heterogeneous domains, often yielding low inlier ratios and unstable performance.

Recent methods~\cite{Li2021DeepI2PIC,Li20232D3DMATR2M,Ren2022CorrI2PDI,Zhou2023DifferentiableRO,mu2025diff2i2p} instead use detection-free pipelines based on direct cross-modal feature interaction. Many adopt coarse-to-fine strategies~\cite{Li20232D3DMATR2M,mu2025diff2i2p}, first identifying globally consistent regions and then refining them into accurate pixel--point correspondences. Diffusion-based approaches~\cite{wang2023freereg} provide an alternative formulation for extracting cross-modal correspondences, but often incur substantial computational overhead.

Most existing methods improve matching through stronger architectures, feature interaction, or refinement, while optimizing cross-modal representations only implicitly through a high-capacity matcher. Consequently, models may depend heavily on global context and complex interactions, while the underlying pair-level representations remain fragile under noise, low overlap, and ambiguous structures. Matching accuracy alone therefore cannot fully characterize representation quality, motivating explicit supervision of representation robustness and recoverability.

\subsection{Probing and Knowledge Distillation}

Our work is related to probe-based representation analysis and teacher--student learning. Lightweight probes examine what information is encoded in learned representations~\cite{Alain2016UnderstandingIL,Hewitt2019DesigningAI}, while knowledge distillation trains a student to mimic a stronger teacher for compression or improved generalization~\cite{Hinton2015DistillingTK,Romero2014FitNetsHF,Zagoruyko2016PayingMA}. Both assess representation quality through how effectively another learner can utilize it.

TeaMatch instead uses weak learners as training-time regularizers rather than analysis tools or deployable models. Low-capacity and structurally degraded learners probe the recoverability of teacher-produced pair representations, ensuring that student performance reflects intrinsic representation quality rather than model capacity. A patch-disjoint teach/meta split prevents trivial memorization and promotes generalization across local regions. Unlike standard distillation centered on feature or prediction imitation, TeaMatch aligns recoverability with downstream geometric objectives, improving robustness and geometric consistency without affecting inference.

\section{Method}

\begin{figure*}[t]
\vspace{-4pt}
  \centering
  \includegraphics[width=\textwidth]{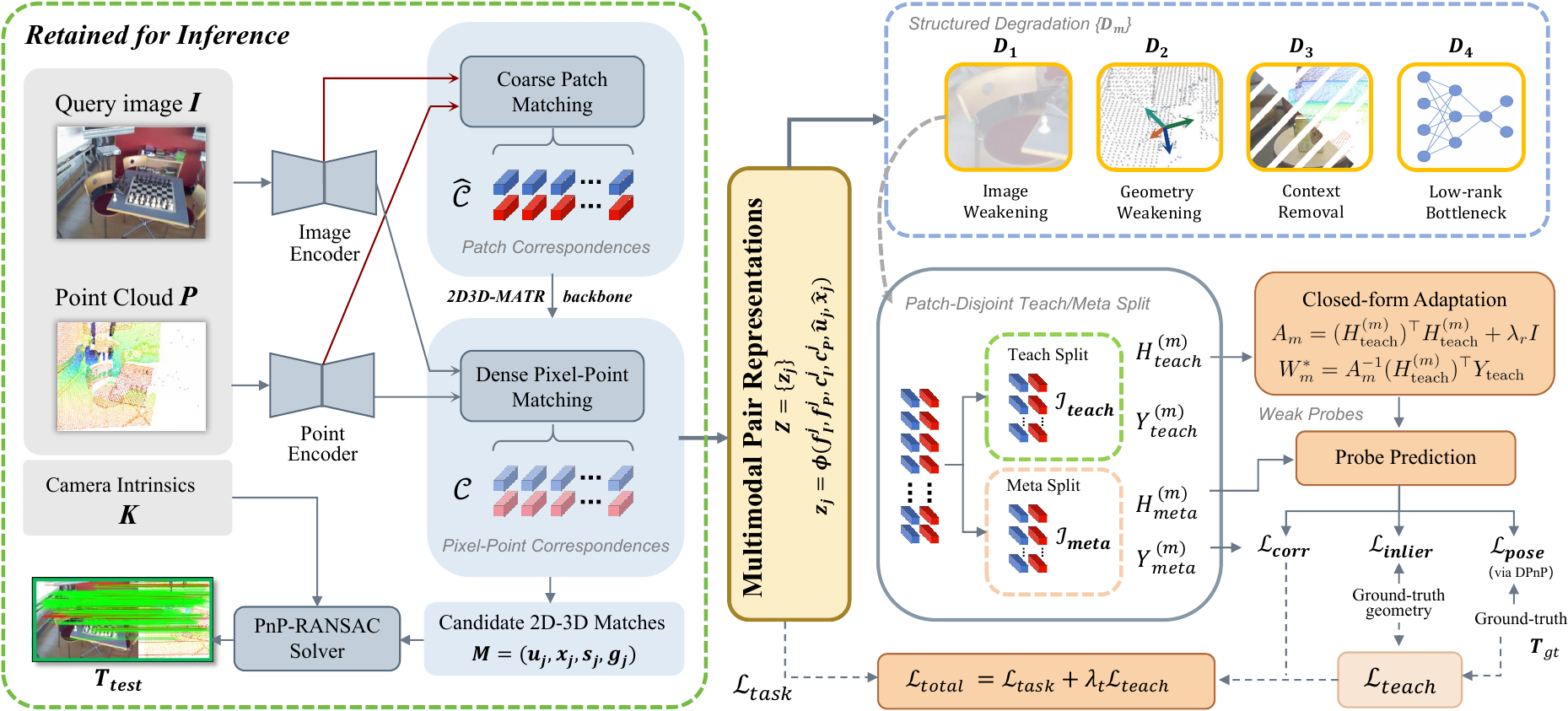}
  \caption{\textbf{Overview of TeaMatch.}
  A detection-free coarse-to-fine teacher first produces candidate 2D--3D correspondences and their pair-level multimodal representations.
  During training, TeaMatch degrades these representations, fits a family of low-capacity weak probes on a \emph{teach split}, and evaluates them on a patch-disjoint \emph{meta split}.
  The resulting teachability signal is imposed at the correspondence, inlierness, and pose levels to regularize the teacher.
  During inference, all probes and teachability branches are removed, and the pipeline reduces exactly to the original teacher matcher followed by a standard pose solver.}
  \label{fig:overview}
\vspace{-8pt}
\end{figure*}

\subsection{Overview}
Given a query image $I \in \mathbb{R}^{H \times W \times 3}$, a scene point cloud
$P=\{x_i\}_{i=1}^{N}$ with $x_i \in \mathbb{R}^3$, and camera intrinsics $K$, the goal of correspondence-based 2D--3D localization is to estimate the camera pose $T=[R|t]\in SE(3)$, where
$R\in SO(3)$ and $t\in\mathbb{R}^3$. A standard pipeline first extracts a set of 2D--3D correspondences
\[
\mathcal{C}=\{(x_k,u_k)\}_{k=1}^{M},
\]
where $x_k$ is a 3D point and $u_k$ is a 2D image location. The pose is then estimated by minimizing the 2D projection error
\begin{equation}
\min_{R,t}
\sum_{(x_k,u_k)\in\mathcal{C}}
\left\|
\pi\!\left(K(Rx_k+t)\right)-u_k
\right\|_2^2,
\end{equation}
where $\pi(\cdot)$ denotes the projection from 3D space to the image plane. This problem can be effectively solved by PnP-RANSAC~\cite{Fischler1981RandomSC,NonameMN}. However, the estimated pose can still be inaccurate when the recovered cross-modal correspondences are unreliable.

We build TeaMatch on top of a standard detection-free coarse-to-fine matcher and treat \emph{teachability} as a training criterion for pair-level multimodal representations. Beyond accurate matching scores, TeaMatch requires that the learned image--point representations remain recoverable under constrained cross-modal reasoning. As shown in Fig.~\ref{fig:overview}, the teacher produces coarse patch correspondences and fine-level candidate matches, which are converted into pair-level representations. During training, we apply structured degradation, fit weak probes on a teach split, and evaluate them on a disjoint meta split to derive teachability losses at the correspondence, inlierness, and pose levels. During inference, all auxiliary components are removed, and the pipeline reduces to the original matcher without additional cost.

\subsection{Teacher and Pair Representation}

We adopt a standard detection-free coarse-to-fine 2D--3D matcher as the teacher.
The image encoder $E_I$ extracts multi-scale visual features from the query image,
while the point-cloud encoder $E_P$ extracts hierarchical geometric features from
the scene point cloud.

In the coarse stage, cross-modal interaction produces a similarity matrix
\begin{equation}
S^c \in \mathbb{R}^{M \times N},
\end{equation}
where each entry measures the affinity between an image patch and a point-cloud patch.
Based on $S^c$, we extract a set of coarse patch correspondences
\begin{equation}
\hat{\mathcal{C}} = \{(a_k, b_k)\}_{k=1}^{K}.
\end{equation}

In the fine stage, local matching is performed within each selected coarse pair,
yielding a set of candidate correspondences
\begin{equation}
\mathcal{M} = \{(u_j, x_j, s_j, g_j)\}_{j=1}^{L},
\end{equation}
where $u_j$ is the 2D image location, $x_j$ is the matched 3D point,
$s_j \in \mathbb{R}$ is the teacher-predicted matching logit, and
$g_j \in \{1, \dots, |\hat{\mathcal{C}}|\}$ denotes the index of the coarse
patch pair from which this candidate originates.

Rather than directly regularizing the final score $s_j$, TeaMatch operates on
the \emph{pair-level multimodal representation} associated with each candidate
correspondence. Specifically, for each candidate match, we construct
\begin{equation}
z_j = \phi(f_j^I, f_j^P, c_j^I, c_j^P, \hat{u}_j, \hat{x}_j) \in \mathbb{R}^{d},
\end{equation}
where $f_j^I$ and $f_j^P$ are the local image and point features,
$c_j^I$ and $c_j^P$ are the corresponding coarse contextual features, and
$\hat{u}_j$ and $\hat{x}_j$ are normalized spatial coordinates.
Here $\phi(\cdot)$ denotes feature concatenation followed by a linear projection.

The teacher is optimized with the standard task loss
\begin{equation}
\mathcal{L}_{\text{task}}
=
\mathcal{L}_c
+
\lambda_f \mathcal{L}_f
+
\lambda_p \mathcal{L}_{\text{pose}}^{T},
\end{equation}
where $\mathcal{L}_c$ and $\mathcal{L}_f$ denote the coarse- and fine-level matching losses,
$\mathcal{L}_{\text{pose}}^{T}$ denotes the teacher pose loss, and $\lambda_f$ and
$\lambda_p$ are hyperparameters that balance the contributions of different loss terms.

\subsection{Teachability via Weak Probes}

TeaMatch tests whether the teacher-produced pair representations remain recoverable
under constrained cross-modal reasoning. To avoid conflating representation quality
with additional model capacity, we introduce a family of weak probes rather than an
auxiliary student network. Each probe operates only on $z_j$, has deliberately limited
capacity, is adapted in closed form, and is removed after training.

\subsubsection{Degradation}

To simulate realistic failure modes in detection-free 2D--3D matching, we define a
family of degradation operators $\{D_m\}_{m\in\mathcal{S}}$ that weaken $z_j$ in
different ways:
\begin{equation}
\tilde{z}_j^{(m)}
=
D_m(z_j)
=
M_m z_j + \eta_j^{(m)},
\end{equation}
where $M_m\in\mathbb{R}^{d\times d}$ is a mode-specific degradation matrix and
$\eta_j^{(m)}$ is optional perturbation noise.

In practice, we use four representative modes:
\begin{itemize}
    \item \textbf{Image weakening:} dropping or masking channels associated with $f_j^I$;
    \item \textbf{Geometry weakening:} perturbing channels associated with $f_j^P$ and/or $\hat{x}_j$;
    \item \textbf{Context removal:} suppressing channels associated with $c_j^I$ and $c_j^P$;
    \item \textbf{Low-rank bottleneck:} projecting $z_j$ to a low-dimensional subspace with $d_b \ll d$.
\end{itemize}

\subsubsection{Probe Feature}

Given a degraded representation, the corresponding probe feature is defined as
\begin{equation}
h_j^{(m)}
=
B_m(\tilde{z}_j^{(m)})
=
W_m^{B}\tilde{z}_j^{(m)},
\end{equation}
where $W_m^{B}\in\mathbb{R}^{d_h\times d}$ is a lightweight mode-specific linear map
and $d_h \ll d$, thereby keeping the probe low-capacity.

\subsubsection{Teach/Meta Split}

A naive probe may achieve overly optimistic performance by exploiting redundancy
within the same local region. To avoid this, TeaMatch partitions candidate
correspondences according to their coarse patch-pair indices $\{g_j\}$ rather than
at the individual-match level.

For each training sample, we randomly split the index set of coarse patch pairs
$\{1,\dots,|\hat{\mathcal{C}}|\}$ into a teach subset $\mathcal{G}_{\text{teach}}$
and a meta subset $\mathcal{G}_{\text{meta}}$ with ratio $\rho:(1-\rho)$. The
candidate matches are then assigned accordingly:
\begin{align}
\mathcal{M}_{\text{teach}}
&=
\left\{
(u_j,x_j,s_j,g_j)\mid
g_j\in \mathcal{G}_{\text{teach}}
\right\},\\
\mathcal{M}_{\text{meta}}
&=
\left\{
(u_j,x_j,s_j,g_j)\mid
g_j\in \mathcal{G}_{\text{meta}}
\right\}.
\end{align}
This patch-disjoint split forces the probe to generalize across different local
regions instead of reusing region-specific statistics.

\subsubsection{Targets}

The weak probes are trained to recover the teacher's soft preference over candidate
correspondences. We define the scalar soft target of each candidate as
\begin{equation}
y_j
=
\sigma(s_j/\tau),
\end{equation}
where $\sigma(\cdot)$ is the sigmoid function and $\tau$ is the temperature.
The teach and meta targets are formed by stacking the corresponding scalar targets
from $\mathcal{M}_{\text{teach}}$ and $\mathcal{M}_{\text{meta}}$, yielding
$Y_{\text{teach}}\in\mathbb{R}^{n_t}$ and $Y_{\text{meta}}\in\mathbb{R}^{n_m}$.

\subsubsection{Closed-Form Fit}

Given probe features on the teach split,
$H_{\text{teach}}^{(m)}\in\mathbb{R}^{n_t\times d_h}$, and the corresponding soft
targets $Y_{\text{teach}}\in\mathbb{R}^{n_t}$, we fit a ridge-regression probe by
solving
\begin{equation}
W_m^*
=
\arg\min_{W\in\mathbb{R}^{d_h}}
\left\|
H_{\text{teach}}^{(m)}W
-
Y_{\text{teach}}
\right\|_2^2
+
\lambda_r \|W\|_2^2.
\end{equation}
Its closed-form solution is
\begin{equation}
W_m^*
=
\left(
(H_{\text{teach}}^{(m)})^\top H_{\text{teach}}^{(m)}
+
\lambda_r I
\right)^{-1}
(H_{\text{teach}}^{(m)})^\top
Y_{\text{teach}}.
\end{equation}
The adapted probe is then evaluated on the meta split:
\begin{equation}
P_{\text{meta}}^{(m)}
=
H_{\text{meta}}^{(m)}W_m^*,
\end{equation}
where $H_{\text{meta}}^{(m)}\in\mathbb{R}^{n_m\times d_h}$.

We fit one probe per degradation mode for each training sample, and recompute the
closed-form solution in every mini-batch. During backpropagation, gradients are
stopped through $W_m^*$ and propagated only through the teacher-produced
representations and the lightweight feature map $B_m(\cdot)$.

\subsection{Teachability Losses}

The probe predictions on the meta split are converted into three teachability
losses. Together, they measure whether the recoverable part of the representation
is consistent with teacher preference, aligned with valid geometry, and useful for
downstream localization.

\subsubsection{Correspondence}

At the correspondence level, the probe is required to recover the teacher's soft
preference on unseen local regions:
\begin{equation}
\mathcal{L}_{\text{corr}}^{(m)}
=
\frac{1}{n_m}
\left\|
P_{\text{meta}}^{(m)}
-
Y_{\text{meta}}
\right\|_2^2.
\end{equation}

\subsubsection{Inlierness}

Recovering teacher preference alone is insufficient, since a probe may match score
patterns that are only weakly aligned with downstream geometry. Let
$\mathcal{I}_{\text{meta}}$ denote the index set of candidate matches in the meta
split, and let $r_j$ denote the geometric residual of the $j$-th candidate under
the ground-truth pose $T_{\mathrm{gt}}$. For each candidate in the meta split, we
define the soft inlier target
\begin{equation}
q_j
=
\exp(-r_j^2/\delta^2),
\end{equation}
where $\delta$ is a scale parameter. Using the probe confidence
\begin{equation}
p_j^{(m)}
=
\sigma(P_{\text{meta},j}^{(m)}),
\end{equation}
we compute the soft inlier ratio
\begin{equation}
\text{SoftIR}^{(m)}
=
\frac{
\sum_{j\in\mathcal{I}_{\text{meta}}} p_j^{(m)} q_j
}{
\sum_{j\in\mathcal{I}_{\text{meta}}} p_j^{(m)} + \epsilon
},
\end{equation}
and define
\begin{equation}
\mathcal{L}_{\text{inlier}}^{(m)}
=
1-\text{SoftIR}^{(m)}.
\end{equation}

\subsubsection{Pose}

Ultimately, pose recovery is the final objective of 2D--3D localization. We
therefore further align teachability with the downstream task by defining a pose
loss on probe-induced correspondences. Specifically, we use the meta-split
correspondences together with their probe confidences to estimate a pose
\begin{equation}
\hat{T}^{(m)}
=
\text{DPnP}
\big(
\{(x_j,u_j,p_j^{(m)})\}_{j\in\mathcal{I}_{\text{meta}}}
\big),
\end{equation}
where $\text{DPnP}(\cdot)$ denotes a differentiable weighted PnP solver. The
resulting pose is supervised against the ground-truth pose:
\begin{equation}
\mathcal{L}_{\text{pose}}^{(m)}
=
\ell_{\text{pose}}(\hat{T}^{(m)},T_{\mathrm{gt}}).
\end{equation}
If a differentiable solver is unavailable, we adopt a stop-gradient pose proxy,
where the PnP solution is treated as a fixed numeric target and gradients are
propagated only through the probe-induced weights prior to pose estimation.

\subsubsection{Overall Loss}

We define the overall teachability objective by averaging over all degradation
modes:
\begin{equation}
\mathcal{L}_{\text{teach}}
=
\frac{1}{|\mathcal{S}|}
\sum_{m\in\mathcal{S}}
\Big(
\beta_1 \mathcal{L}_{\text{corr}}^{(m)}
+
\beta_2 \mathcal{L}_{\text{inlier}}^{(m)}
+
\beta_3 \mathcal{L}_{\text{pose}}^{(m)}
\Big),
\end{equation}
where $\beta_1$, $\beta_2$, and $\beta_3$ are balancing weights for the three
teachability losses. The final training objective is
\begin{equation}
\mathcal{L}
=
\mathcal{L}_{\text{task}}
+
\lambda_t \mathcal{L}_{\text{teach}},
\end{equation}
where $\lambda_t$ controls the contribution of the teachability objective.

\subsection{Training and Inference}
TeaMatch is a pure training-time framework that operates on teacher-generated pair representations and introduces weak probes as auxiliary constraints to regularize multimodal representation quality, without modifying the original matching objective. During inference, all degradation operators, weak probes, and teachability branches are removed, and the pipeline reduces exactly to the original teacher matcher followed by a standard PnP-RANSAC\cite{Fischler1981RandomSC,NonameMN} solver, incurring no additional inference overhead. Consequently, the performance gains of TeaMatch stem entirely from improved training-time representation learning.

\section{Experiments}
\subsection{Implementation Details}

\noindent\textbf{Network architecture.}
TeaMatch is built upon a detection-free coarse-to-fine 2D--3D matching pipeline. We adopt the same backbone configuration as strong prior methods such as 2D3D-MATR\cite{Li20232D3DMATR2M} and Diff2I2P\cite{mu2025diff2i2p}, using a 4-stage ResNet-FPN\cite{He2015DeepRL} as the image encoder and a 4-stage KPConv-FPN\cite{Thomas2019KPConvFA} as the point cloud encoder. The input image resolution is $480 \times 640$, with the coarsest feature map downsampled to $60 \times 80$. The point cloud is voxelized with an initial voxel size of 2.5\,cm, which is doubled at each stage. TeaMatch introduces teachability regularization only during training, while keeping the teacher architecture unchanged at inference time.

\noindent\textbf{Datasets.}
We evaluate TeaMatch on two widely used benchmarks: 7-Scenes~\cite{Glocker2013RealtimeRC} and RGB-D Scenes V2~\cite{Lai2014UnsupervisedFL}. For 7-Scenes, we follow prior work to construct image--point-cloud pairs with at least 50\% overlap and use the official train/validation/test split to evaluate generalization to unseen viewpoints. For RGB-D Scenes V2, we construct pairs with at least 30\% overlap and split them into training, validation, and testing sets across different scenes to evaluate cross-scene generalization. Our data preparation follows previous methods\cite{Li20232D3DMATR2M} for fair comparison.

\noindent\textbf{Metrics.}
We adopt standard evaluation metrics for 2D--3D matching, including Inlier Ratio (IR), Feature Matching Recall (FMR), Registration Recall (RR), and Patch Inlier Ratio (PIR). IR measures the percentage of correct correspondences within a 3D distance threshold of 5\,cm. FMR measures the proportion of image--point-cloud pairs whose inlier ratio exceeds a threshold of 10\%. RR measures the percentage of pairs whose pose error is below 10\,cm. PIR evaluates the correctness of coarse patch correspondences.

\noindent\textbf{Baselines.}
We compare TeaMatch with representative 2D--3D matching methods, including FCGF-2D3D~\cite{Choy2019FullyCG}, Predator-2D3D~\cite{Huang2020PREDATORRO}, P2-Net~\cite{Wang2021P2NetJD}, 2D3D-MATR~\cite{Li20232D3DMATR2M}, and Diff2I2P~\cite{mu2025diff2i2p}. We follow the standard backbone settings and evaluation protocols used in prior work whenever applicable. Since TeaMatch is applied as a training-time regularization on top of the same teacher architecture, any performance gain is attributable to improved representation learning rather than additional inference modules.

\begin{table}[h]
\caption{Registration results of TeaMatch and baselines on 7-Scenes. The best result for each metric is shown in \textbf{bold}.}
\label{tab:1}
\centering
\footnotesize
\setlength{\tabcolsep}{3.6pt}   
\renewcommand{\arraystretch}{1.1}  
\begin{tabular}{lcccccccc}
\toprule
Model & Chess & Fire & Heads & Office & Pump & Kitchen & Stairs & Mean \\
\midrule
\multicolumn{9}{c}{Inlier Ratio $\uparrow$} \\
\midrule
FCGF\cite{Choy2019FullyCG}            & 34.2 & 32.8 & 14.8 & 26.0 & 23.3 & 22.5 &  6.0 & 22.8 \\
P2Net\cite{Wang2021P2NetJD}           & 55.2 & 46.7 & 13.0 & 36.2 & 32.0 & 32.8 &  5.8 & 31.7 \\
Predator\cite{Huang2020PREDATORRO}        & 34.7 & 33.8 & 16.6 & 25.9 & 23.1 & 22.2 &  7.5 & 23.4 \\
MATR\cite{Li20232D3DMATR2M}            & 72.1 & 66.0 & 31.3 & 60.7 & 50.2 & 52.5 & 18.1 & 50.1 \\
Diff$^2$I2P\cite{mu2025diff2i2p}     & 74.1 & 68.8 & 39.2 & 65.6 & 52.1 & 54.2 & 18.1 & 53.2 \\
\textbf{TeaMatch} & \textbf{76.3} & \textbf{70.8} & \textbf{42.2} & \textbf{66.7} & \textbf{52.4} & \textbf{55.1} & \textbf{18.4} & \textbf{54.6} \\
\midrule
\multicolumn{9}{c}{Feature Matching Recall $\uparrow$} \\
\midrule
FCGF\cite{Choy2019FullyCG}             &  99.7 &  98.2 &  69.9 &  97.1 &  83.0 &  87.7 & 16.2 & 78.8 \\
P2Net\cite{Wang2021P2NetJD}           & 100.0 &  99.3 &  58.9 &  99.1 &  87.2 &  92.2 & 16.2 & 79.0 \\
Predator\cite{Huang2020PREDATORRO}        &  91.3 &  95.1 &  76.7 &  88.6 &  79.2 &  80.6 & 31.1 & 77.5 \\
MATR\cite{Li20232D3DMATR2M}             & 100.0 &  99.6 &  98.6 & 100.0 &  92.4 &  95.9 & 58.1 & 92.1 \\
Diff$^2$I2P\cite{mu2025diff2i2p}     & 100.0 & \textbf{100.0} & 100.0 & 100.0 & 93.4 & 96.2 & 55.4 & 92.2 \\
\textbf{TeaMatch} & \textbf{100.0} & 99.8 & \textbf{100.0} & \textbf{100.0} & \textbf{94.1} & \textbf{96.7} & \textbf{59.2} & \textbf{92.9} \\
\midrule
\multicolumn{9}{c}{Registration Recall $\uparrow$} \\
\midrule
FCGF\cite{Choy2019FullyCG}             & 89.5 & 79.7 & 19.2 & 85.9 & 69.4 & 79.0 &  6.8 & 61.4 \\
P2Net\cite{Wang2021P2NetJD}           & 96.9 & 86.5 & 20.5 & 91.7 & 75.3 & 85.2 &  4.1 & 65.7 \\
Predator\cite{Huang2020PREDATORRO}        & 69.6 & 60.7 & 17.8 & 62.9 & 56.2 & 62.6 &  9.5 & 48.5 \\
MATR\cite{Li20232D3DMATR2M}             & 96.9 & 90.7 & 52.1 & 95.5 & 80.9 & 86.1 & 28.4 & 75.8 \\
Diff$^2$I2P\cite{mu2025diff2i2p}     & \textbf{99.0} & 95.6 & 74.0 & \textbf{98.9} & 86.8 & 90.2 & 36.5 & 83.0 \\
\textbf{TeaMatch} & 98.3 & \textbf{96.1} & \textbf{79.9} & \textbf{98.9} & \textbf{87.3} & \textbf{92.2} & \textbf{43.6} & \textbf{85.2} \\
\bottomrule
\end{tabular}
\end{table}

\begin{figure*}[h]
\vspace{-4pt}
  \centering
  \includegraphics[width=\textwidth]{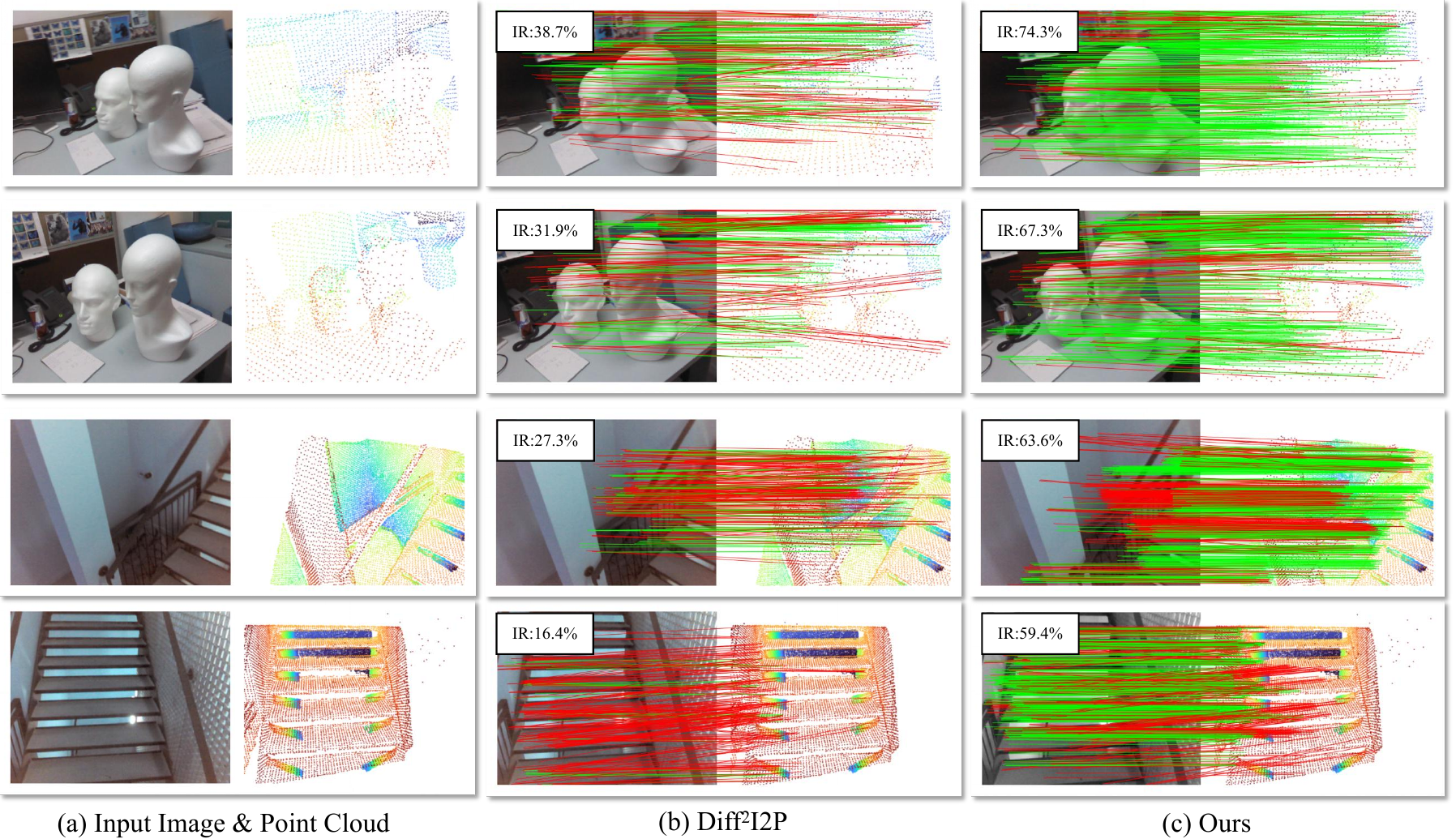}
  \caption{
  \textbf{Qualitative results on the 7-Scenes dataset. The red lines indicate erroneous correspondences (3D distance greater than 5 cm), while the green lines represent correct correspondences.
  }
  \label{fig:3}}
\end{figure*}

\subsection{Evaluations on 7-Scenes}

\noindent\textbf{Quantitative results.}
We evaluate TeaMatch on the 7-Scenes\cite{Glocker2013RealtimeRC} dataset and report the results in Tab.~\ref{tab:1}. For IR, TeaMatch consistently outperforms previous methods, achieving the best mean IR of 54.6\%, indicating more accurate and reliable pixel--point correspondences. This improvement further leads to stronger FMR, where TeaMatch achieves the highest mean FMR of 92.9\% and reaches near-perfect matching performance on multiple scenes. Regarding the most critical metric RR, TeaMatch demonstrates clear and consistent improvements, achieving an average RR of 85.2\%, surpassing 2D3D-MATR\cite{Li20232D3DMATR2M} by 9.4\% and Diff2I2P\cite{mu2025diff2i2p} by 2.2\%. The gains are particularly notable on challenging scenes such as \emph{Heads} and \emph{Stairs}, where TeaMatch shows stronger robustness under local ambiguity and repetitive structures. These results demonstrate that improving representation teachability during training leads to more reliable correspondences and better downstream pose estimation.

\noindent\textbf{Qualitative results.}
Fig.~\ref{fig:3} visualizes the correspondences produced by Diff$^2$I2P and TeaMatch on challenging scenes from the 7-Scenes dataset. Column (a) shows the input image and point cloud pairs, while columns (b) and (c) present the results from Diff$^2$I2P and TeaMatch, respectively. The first two rows correspond to the \emph{Heads} scene, which exhibits strong geometric ambiguity and limited texture cues. In these cases, Diff$^2$I2P produces cluttered matches with many outliers, yielding low inlier ratios around 31\%--39\%, whereas TeaMatch generates cleaner and more localized correspondences with significantly higher inlier ratios of 67\%--74\%. The last two rows show the \emph{Stairs} scene, which is particularly challenging due to repetitive structures and weak geometric distinctiveness. Here, the baseline suffers from severe mismatches (IR as low as 16\%--27\%), while TeaMatch effectively suppresses outliers and improves spatial consistency, achieving much higher inlier ratios of 59\%--64\%. These results demonstrate that TeaMatch produces more reliable correspondences under challenging conditions by learning more robust and geometrically consistent representations.

\begin{figure*}[t]
\vspace{-4pt}
  \centering
  \includegraphics[width=\textwidth]{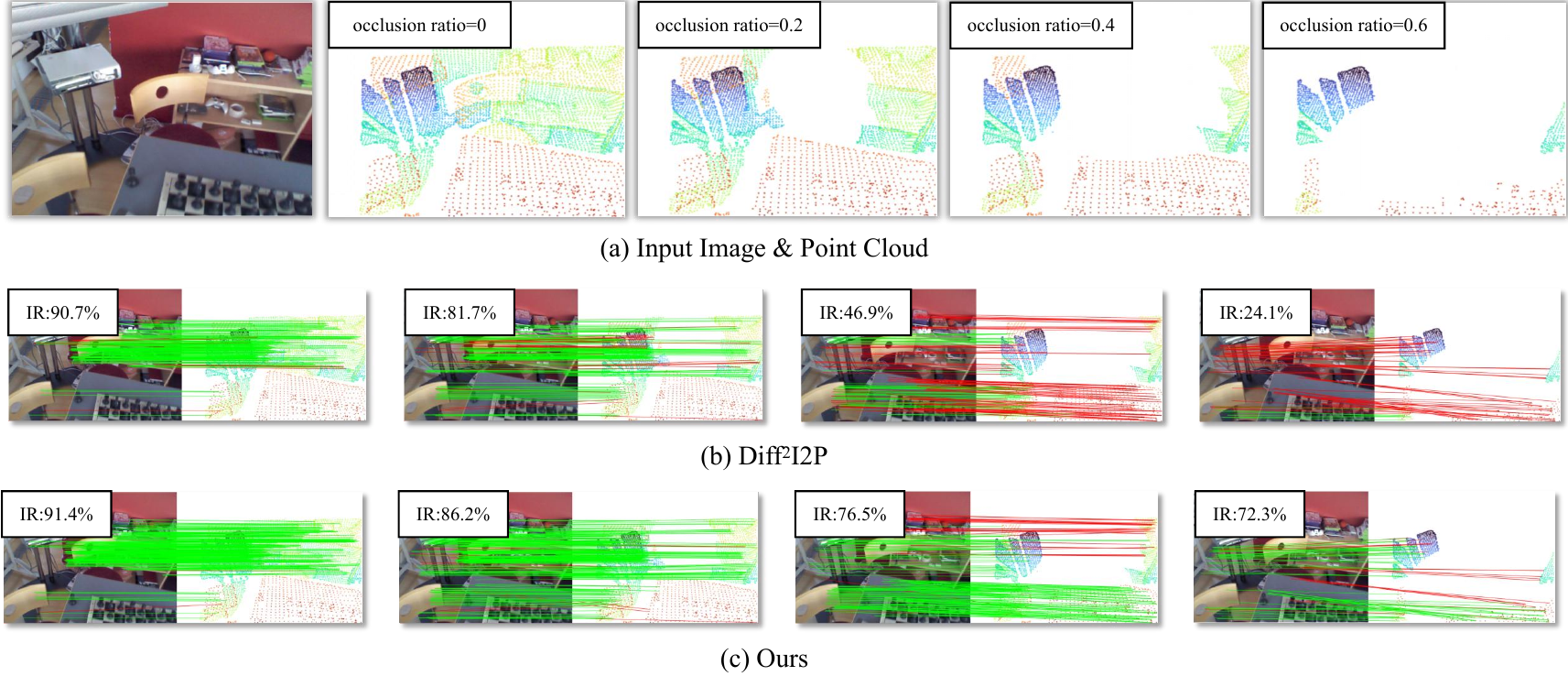}
  \caption{
\textbf{Robustness under local point cloud degradation.}
From left to right, the occlusion ratio increases from $0$ to $0.6$, indicating progressively more severe local dropout in the point cloud.
(a) Input image and degraded point cloud.
(b) Correspondences produced by Diff$^2$I2P.
(c) Correspondences produced by TeaMatch.
Green and red lines denote correct and incorrect matches, respectively, and the inlier ratio (IR) is reported for each case.
TeaMatch consistently achieves higher IR and remains robust under severe structural degradation.
}
\label{fig:4}
\vspace{-8pt}
\end{figure*}

\subsection{Evaluations on RGB-D Scenes V2}

\noindent\textbf{Quantitative results.}
We further evaluate TeaMatch on RGB-D Scenes V2\cite{Lai2014UnsupervisedFL}, which emphasizes generalization to unseen scenes. The results are reported in Tab.~\ref{tab:2}. TeaMatch achieves the best performance across all metrics, reaching 63.1\% in RR and 78.3\% in FMR. Compared with Diff$^2$I2P\cite{mu2025diff2i2p}, our method brings consistent improvements across all metrics, with a gain of about 2.6\% in RR. The improvement over 2D3D-MATR\cite{Li20232D3DMATR2M} is more pronounced, especially in PIR and RR. These results indicate that TeaMatch improves both coarse and fine correspondence estimation, leading to more stable matching and more robust pose estimation under cross-scene distribution shifts.

\begin{table}[h]
\caption{Registration results of TeaMatch and baselines on RGB-D Scenes V2. The best result for each metric is shown in \textbf{bold}.}
\label{tab:2}
\centering
\footnotesize
\setlength{\tabcolsep}{11.5pt}   
\renewcommand{\arraystretch}{1.1}  
\begin{tabular}{l|cccc}
\toprule
Method & PIR (\%)$\uparrow$ & IR (\%)$\uparrow$ & FMR (\%)$\uparrow$ & RR (\%)$\uparrow$ \\
\midrule
FCGF\cite{Choy2019FullyCG}       & 20.1 & 10.3 & 29.2 & 32.5 \\
P2-Net\cite{Wang2021P2NetJD}           & 30.4 & 14.5 & 63.7 & 41.7 \\
Predator\cite{Huang2020PREDATORRO}    & 32.6 & 15.8 & 68.1 & 33.6 \\
MATR\cite{Li20232D3DMATR2M}        & 57.6 & 36.3 & 76.0 & 56.9 \\
Diff$^2$I2P\cite{mu2025diff2i2p}      & 60.8 & 36.9 & 77.1 & 60.5 \\
\textbf{TeaMatch} & \textbf{61.3} & \textbf{37.6} & \textbf{78.3} & \textbf{63.1} \\
\bottomrule
\end{tabular}
\end{table}

\subsection{Ablation Studies}

We conduct extensive ablation studies to validate the contribution of each component in TeaMatch. Unless otherwise specified, all experiments are performed on 7-Scenes, and we report IR, FMR, and RR for evaluation.

\noindent\textbf{Overall effect of TeaMatch.}
We evaluate the overall impact of TeaMatch by comparing the teacher baseline with and without our framework. As shown in Tab.~\ref{tab:3}, TeaMatch consistently improves all evaluation metrics, with IR increasing from 50.1\% to 54.6\% and RR from 75.8\% to 85.2\%. This clear improvement verifies that TeaMatch provides a strong overall performance gain. Importantly, since all teachability-related components are removed during inference, these gains are achieved purely through improved training-time representation learning without introducing additional computational overhead.

\begin{table}[h]
\caption{Overall effect of TeaMatch on 7-Scenes. TeaMatch consistently improves all evaluation metrics over the teacher baseline without modifying the inference pipeline.}
\centering
\footnotesize
\setlength{\tabcolsep}{13pt}
\renewcommand{\arraystretch}{1.2}
\begin{tabular}{lrrrr}
\toprule
Model & PIR $\uparrow$ & IR $\uparrow$ & FMR $\uparrow$ & RR $\uparrow$ \\
\midrule
Teacher baseline & 83.4  & 50.1 & 92.1  & 75.8 \\
TeaMatch (ours)  & \textbf{86.9} & \textbf{54.6} & \textbf{92.9} & \textbf{85.2} \\
\bottomrule
\end{tabular}
\label{tab:3}
\end{table}

\noindent\textbf{Effect of teachability losses.}
We further analyze the contribution of each teachability loss. As shown in Tab.~\ref{tab:loss}, using only the correspondence loss $\mathcal{L}_{corr}$ already improves IR from 50.1\% to 51.3\%, indicating that enforcing recoverability stabilizes pair-level representations. Adding the inlierness loss $\mathcal{L}_{inlier}$ further improves IR and FMR by encouraging geometry-aware discrimination of correspondences. Incorporating the pose loss $\mathcal{L}_{pose}$ brings a larger gain in RR (from 79.6\% to 83.0\%), as it directly aligns the training objective with downstream pose estimation. The full model achieves the best performance, demonstrating that the three losses provide complementary supervision.

\begin{table}[h]
\caption{Ablation study of teachability losses on 7-Scenes. $\mathcal{L}_{corr}$, $\mathcal{L}_{inlier}$, and $\mathcal{L}_{pose}$ denote the correspondence, inlierness, and pose losses, respectively.}
\centering
\footnotesize
\setlength{\tabcolsep}{4pt}
\renewcommand{\arraystretch}{1.1}
\begin{tabular}{lccc cccc}
\toprule
Type & $\mathcal{L}_{corr}$ & $\mathcal{L}_{inlier}$ & $\mathcal{L}_{pose}$ 
& PIR $\uparrow$ & IR $\uparrow$ & FMR $\uparrow$ & RR $\uparrow$ \\
\midrule
(a) baseline        &  &  &  & 83.4 & 50.1 & 92.1 & 75.8 \\
(b) + corr          & $\checkmark$ &  &  & 84.9 & 51.3 & 92.3 & 79.6 \\
(c) + corr + inlier & $\checkmark$ & $\checkmark$ &  & 85.8 & 53.6 & 92.6 & 82.1 \\
(d) + corr + pose   & $\checkmark$ &  & $\checkmark$ & 85.5 & 52.7 & 92.5 & 83.0 \\
(e) full model      & $\checkmark$ & $\checkmark$ & $\checkmark$ & \textbf{86.9} & \textbf{54.6} & \textbf{92.9} & \textbf{85.2} \\
\bottomrule
\end{tabular}
\label{tab:loss}
\end{table}

\noindent\textbf{Effect of teach/meta split.}
We compare the proposed patch-disjoint teach/meta split with two alternatives: no split and random match-level split. As shown in Tab.~\ref{tab:5}, the patch-disjoint split achieves the best performance, reaching 54.6\% IR and 85.2\% RR. In contrast, using no split or random splitting leads to inferior results. This is because, without proper partitioning, the probe can exploit local redundancy within the same region, resulting in overly optimistic supervision. The patch-disjoint split instead enforces generalization across different coarse regions, providing a more effective recoverability constraint.

\begin{table}[h]
\caption{Ablation study of teach/meta split strategies. The patch-disjoint split yields the best performance.}
\centering
\footnotesize
\setlength{\tabcolsep}{10pt}
\renewcommand{\arraystretch}{1.1}
\begin{tabular}{lcccc}
\toprule
Split strategy & PIR $\uparrow$ & IR $\uparrow$ & FMR $\uparrow$ & RR $\uparrow$ \\
\midrule
No split            & 84.7 & 53.0 & 92.3 & 79.6 \\
Random split        & 85.8 & 53.9 & 92.6 & 82.4 \\
Patch-disjoint (ours) & \textbf{86.9} & \textbf{54.6} & \textbf{92.9} & \textbf{85.2} \\
\bottomrule
\end{tabular}
\label{tab:5}
\end{table}

\noindent\textbf{Effect of degradation modes.}
We analyze the impact of different degradation strategies, including image weakening, geometry weakening, context removal, and low-rank bottleneck projection. As shown in Tab.~\ref{tab:6}, all individual degradation modes improve performance over the baseline, while combining them yields the best results. In particular, the full model achieves the highest IR (54.6\%) and RR (85.2\%), outperforming all single-mode variants. This indicates that diverse degradation patterns are complementary and jointly encourage more robust cross-modal representation learning.

\begin{table}[h]
\caption{Ablation study of degradation modes. Combining multiple degradation strategies achieves the best performance.}
\centering
\footnotesize
\setlength{\tabcolsep}{11pt}
\renewcommand{\arraystretch}{1.1}
\begin{tabular}{lcccc}
\toprule
Degradation mode & PIR $\uparrow$ & IR $\uparrow$ & FMR $\uparrow$ & RR $\uparrow$ \\
\midrule
None                  & 83.7 & 51.9 & 92.2 & 76.4 \\
Image                 & 85.1 & 53.2 & 92.5 & 80.4 \\
Geometry              & 84.9 & 53.7 & 92.4 & 81.0 \\
Context               & 85.3 & 52.8 & 92.3 & 79.8 \\
Low-rank              & 85.7 & 53.5 & 92.6 & 82.1 \\
All (ours)            & \textbf{86.9} & \textbf{54.6} & \textbf{92.9} & \textbf{85.2} \\
\bottomrule
\end{tabular}
\label{tab:6}
\end{table}

\noindent\textbf{Robustness under local point cloud degradation.}
We further evaluate robustness under controlled local point cloud degradation by randomly removing local structures with increasing occlusion ratio. As shown in Fig.~\ref{fig:4} and Tab.~\ref{tab:7}, the performance of all methods degrades as the occlusion ratio increases. However, TeaMatch consistently maintains higher inlier ratios and more reliable correspondences than Diff$^2$I2P\cite{mu2025diff2i2p} across all degradation levels. In particular, under severe degradation (occlusion ratio = 0.6), Diff$^2$I2P\cite{mu2025diff2i2p} drops to 23.5\% IR, while TeaMatch still achieves 44.2\%, preserving a substantially larger portion of correct matches. This consistent gap demonstrates that TeaMatch degrades more gracefully and remains robust under strong structural corruption.

\begin{table}[h]
\caption{Inlier ratio under different occlusion ratios. TeaMatch degrades more gracefully than baseline methods.}
\centering
\footnotesize
\setlength{\tabcolsep}{14pt}
\renewcommand{\arraystretch}{1.1}
\begin{tabular}{lcccc}
\toprule
Method & 0 & 0.2 & 0.4 & 0.6 \\
\midrule
Diff$^2$I2P\cite{mu2025diff2i2p} & 53.2 & 48.2 & 31.5 & 23.5 \\
TeaMatch (ours) & \textbf{54.6} & \textbf{53.3} & \textbf{47.7} & \textbf{44.2} \\
\bottomrule
\end{tabular}
\label{tab:7}
\end{table}

\noindent\textbf{Efficiency analysis.}
We analyze the efficiency of TeaMatch in terms of inference time, model size, and memory consumption. As shown in Tab.~\ref{tab:9}, TeaMatch achieves improved registration performance while maintaining the same level of efficiency as the teacher baseline. Notably, compared with diffusion-based methods such as FreeReg, which require expensive inference (9.634\,s and 14.2\,GB VRAM), TeaMatch introduces no additional computational overhead at test time. This is because all teachability-related components are used only during training and removed during inference, resulting in an efficient and practical pipeline.

\begin{table}[h]
\caption{Comparison of performance and efficiency. TeaMatch achieves improved accuracy while maintaining efficient inference.}
\centering
\footnotesize
\setlength{\tabcolsep}{7pt}
\renewcommand{\arraystretch}{1.1}
\begin{tabular}{lcccc}
\toprule
Method & RR (\%)$\uparrow$ & Time (s)$\downarrow$ & Size (MB)$\downarrow$ & VRAM (GB)$\downarrow$ \\
\midrule
2D3D-MATR\cite{Li20232D3DMATR2M} & 75.8 & 0.072 & 118.62 & 3.1 \\
Diff$^2$I2P\cite{mu2025diff2i2p} & 83.0 & 0.074 & 118.75 & 3.1 \\
FreeReg\cite{wang2023freereg} & 71.2 & 9.634 & - & 14.2 \\
TeaMatch (ours) & \textbf{85.2} & 0.075 & 118.71 & 3.1 \\
\bottomrule
\end{tabular}
\label{tab:9}
\vspace{-8pt}
\end{table}

\section{Conclusion}

We presented TeaMatch, a teachability-driven training framework for 2D--3D matching that regularizes whether pair-level representations remain recoverable under degraded inputs and constrained reasoning. Without modifying the inference pipeline, TeaMatch consistently improves strong baselines on 7-Scenes and RGB-D Scenes V2, with notable gains under challenging conditions such as repetitive structures and severe local degradation. These results demonstrate that enforcing representation recoverability is an effective way to learn more robust cross-modal representations.

\begin{acks}
This work was supported by the National Natural Science Foundation of China (62373164), the Natural Science Foundation of Shandong Province (ZR2025QC2246Z), the Taishan Scholar Foundation of Shandong Province (tsqn202507271), the Central Government Guides Local Program (YDZX2024075), and the Taishan Experts Program (tscy20241154).
\end{acks}



\bibliographystyle{ACM-Reference-Format}
\balance
\bibliography{main/sample-base}










\end{document}